\documentclass[runningheads]{llncs}
\usepackage[T1]{fontenc}
\usepackage{graphicx}

\usepackage{amsmath,amsfonts,booktabs,pgfplots,multicol,multirow,tikz}
\usepackage[table]{xcolor}
\usepackage[colorlinks=true, allcolors=black]{hyperref}

\definecolor{oursfull}{gray}{0.90}
\definecolor{oursbase}{gray}{0.95}

\newcommand{\best}[1]{\textbf{#1}}
\newcommand{\runnerup}[1]{\underline{#1}}

\definecolor{blue}{HTML}{1F77B4}
\definecolor{orange}{HTML}{FF7F0E}
\definecolor{green}{HTML}{2CA02C}
\definecolor{upthree}{HTML}{BDD7EE} 
\definecolor{uptwo}{HTML}{DDEBF7}   
\definecolor{upone}{HTML}{F2F8FD}   

\definecolor{downthree}{HTML}{F8CBAD} 
\definecolor{downtwo}{HTML}{FCE4D6}   
\definecolor{downone}{HTML}{FEF2EB}   

\pgfplotsset{compat=1.18} 

\begin{document}
\title{FACTUM: Mechanistic Detection of Citation Hallucination in Long-Form RAG}
\titlerunning{FACTUM: Citation Hallucination Detection}
%
%
\author{Maxime Dassen\inst{1}\orcidID{0009-0000-1046-5416} \and
Rebecca Kotula \inst{2}\orcidID{0009-0006-1337-4919} \and
Kenton Murray \inst{3}\orcidID{0000-0002-5628-1003} \and 
Andrew Yates \inst{3}\orcidID{0000-0002-5970-880X} \and 
Dawn Lawrie \inst{3}\orcidID{0000-0001-7347-7086} \and 
Efsun Kayi \inst{3}\orcidID{0000-0002-2357-9807} \and 
James Mayfield \inst{3}\orcidID{0000-0003-3866-3013} \and 
Kevin Duh \inst{3}\orcidID{0000-0001-8107-4383} }

\authorrunning{M. Dassen et al.}

\institute{University of Amsterdam, Amsterdam, The Netherlands \\
\email{m.s.w.dassen@uva.nl} \and
Department of Defense, Washington D.C., USA \\
\email{becca.kotula@gmail.com} \and
HLTCOE, Johns Hopkins University, Baltimore, Maryland, USA \\
\email{\{kenton, andrew.yates, lawrie, mayfield, kduh1\}@jhu.edu} \and
Applied Physics Laboratory, Johns Hopkins University, Laurel, Maryland, USA \\
\email{efsun.kayi@jhuapl.edu}}
\authorrunning{M. Dassen et al.}
%
%
\maketitle 
\begin{abstract}
Retrieval-Augmented Generation (RAG) models are critically undermined by citation hallucinations, a deceptive failure where a model cites a source that fails to support its claim. While existing work attributes hallucination to a simple over-reliance on parametric knowledge, we reframe this failure as an evolving, scale-dependent coordination failure between the Attention (\textit{reading}) and Feed-Forward Network (\textit{recalling}) pathways. We introduce \textbf{FACTUM} (\textbf{F}ramework for \textbf{A}ttesting \textbf{C}itation \textbf{T}rustworthiness via \textbf{U}nderlying \textbf{M}echanisms), a framework of four mechanistic scores: Contextual Alignment (CAS), Attention Sink Usage (BAS), Parametric Force (PFS), and Pathway Alignment (PAS). Our analysis reveals that correct citations are consistently marked by higher parametric force (PFS) and greater use of the attention sink (BAS) for information synthesis. Crucially, we find that ``one-size-fits-all'' theories are insufficient as the signature of correctness evolves with scale: while the 3B model relies on high pathway alignment (PAS), our best-performing 8B detector identifies a shift toward a specialized strategy where pathways provide distinct, orthogonal information. By capturing this complex interplay, FACTUM outperforms state-of-the-art baselines by up to 37.5\% in AUC. Our results demonstrate that high parametric force is constructive when successfully coordinated with the Attention pathway, paving the way for more nuanced and reliable RAG systems.
\end{abstract}


\section{Introduction}

\begin{figure}[htb!]
    \centering
    \includegraphics[width=0.82\textwidth]{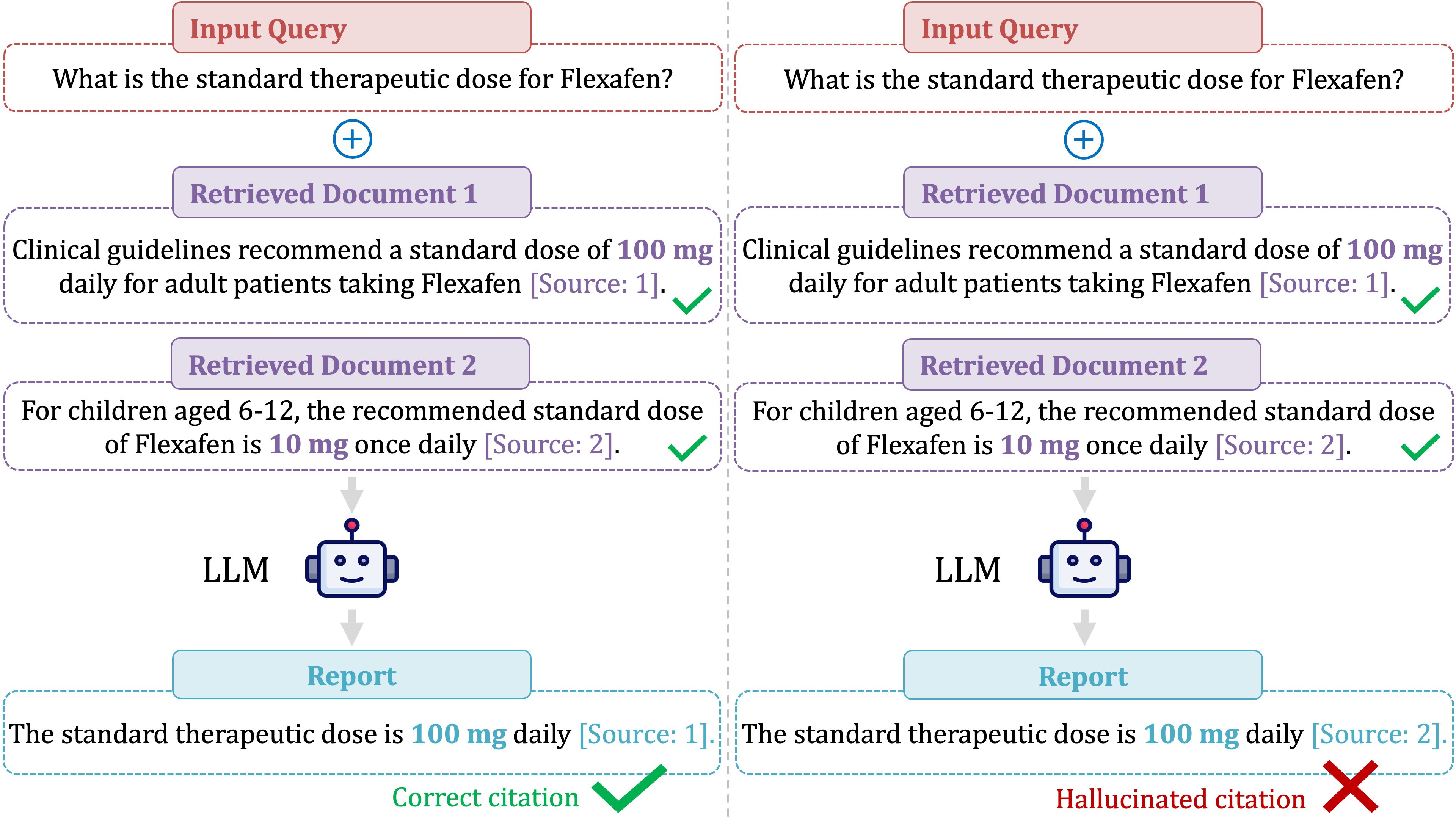} 
    \footnotesize
    \caption{An example demonstrating how citation hallucination makes factually correct information dangerously unreliable. Both scenarios receive the same query and retrieved documents. Left (Correct Citation): The model correctly cites its source `[Source: 1]', making the claim verifiable. Right (Hallucinated Citation): The model generates the same correct fact but falsely attributes it to `[Source: 2]', pointing the user to conflicting information and making the claim unverifiable.}
    \label{fig:main_example}
\end{figure}

Retrieval-Augmented Generation (RAG) systems are designed to improve the accuracy of Large Language Models (LLMs) by integrating relevant information retrieved from external knowledge sources \cite{lewis2021retrievalaugmentedgenerationknowledgeintensivenlp,ayala2024}. However, even with access to accurate information, these models can still produce statements that misrepresent or contradict the provided sources, a phenomenon known as RAG Hallucination \cite{niu2024ragtruthhallucinationcorpusdeveloping,sun2025redeepdetectinghallucinationretrievalaugmented}. 
While prior work leverages external context and parametric knowledge to detect general RAG hallucinations \cite{sun2025redeepdetectinghallucinationretrievalaugmented,tan2025dynamicparametricretrievalaugmented,li2025ragddroptimizingretrievalaugmentedgeneration}, the critical issue of citation hallucination detection has been largely unaddressed. We argue that detecting citation hallucinations in RAG is a distinct problem that deserves its own study.

Citation hallucination is the misattribution of information to an incorrect or fabricated source \cite{magesh2024hallucinationfreeassessingreliabilityleading}. Figure~\ref{fig:main_example} demonstrates how a hallucinated citation makes even factually correct information dangerously unreliable. The hallucination of citations is particularly problematic because the presence of citations significantly increases a reader's trust in the presented information \cite{ding2025citationstrustllmgenerated,shakespeare}. Moreover, studies show that users are influenced by the sheer number of citations, even when the cited content fails to support the claims that were made \cite{miroyan2025searcharenaanalyzingsearchaugmented}. Consequently, a model that generates hallucinated citations breaks the fundamental promise of verifiability that a reference is meant to provide \cite{ding2025citationstrustllmgenerated}.

This becomes particularly challenging in long-form RAG, where LLMs are tasked with generating extended, multi-paragraph reports from vast document collections\footnote{Modern long-context models can process context windows from 32k to 2M tokens, equivalent to hundreds or thousands of pages of text \cite{li2025longcontext}.}. In this setting, the model must maintain correct attribution throughout a long output, introducing the risk of attribution drift: the longer the model writes, the more likely contextual signals regarding source attribution dissipate, leading to misattribution \cite{patel-etal-2024-towards}. To identify the internal signatures that characterize hallucination of citations, our work moves beyond black-box analysis to investigate the measurable mechanistic signals within the model's internal states.

To locate these signals, we ground our approach in the transformer's architecture, where each layer processes information through two distinct submodules with fundamentally different roles. The attention mechanism serves as the pathway to retrieve information from the external context, processing the source documents provided \cite{lietal2019information,daietal2022knowledge}. In contrast, Feed-Forward networks (FFNs) are the primary pathway to apply the stored parametric knowledge of the model \cite{huang2025parammutesuppressingknowledgecriticalffns,NEURIPS20248936fa16}. This architectural division allows us to conceptualize them as two primary information pathways: what the model \textit{reads} and what it \textit{recalls}. By decoupling their contributions to the model's residual stream, which is the shared communication channel that sub-layers both read from and write updates to, we can directly probe the interplay between what the model \textit{reads} from the external context and what it \textit{recalls} from parametric memory \cite{elhage2021mathematical,mak2025residual}. 

Prior mechanistic analyses of general RAG hallucination argued that hallucinations predominantly arise from an over-reliance on the model's parametric knowledge \cite{sun2025redeepdetectinghallucinationretrievalaugmented}. We argue that for citation attribution, this view is incomplete. The core failure is not a simple over-reliance on parametric knowledge, but a systemic breakdown in how the model utilizes its internal mechanisms. Detecting such failures reliably necessitates evaluating the model's mechanistic state during citation generation. To this end, we ask a series of interconnected questions: Regarding what the model \textit{reads} via its Attention pathway: How strongly is the citation grounded in the source documents, and how is that information synthesized across the model's layers? Regarding what the model \textit{recalls} from its FFNs: How forcefully does it apply its own parametric knowledge? Finally, how do these two pathways interact? Are they working in concert, or are they contributing distinct, orthogonal information?

To answer these questions, our contributions are threefold: (1) we introduce \textbf{FACTUM} (\textbf{F}ramework for \textbf{A}ttesting \textbf{C}itation \textbf{T}rustworthiness via \textbf{U}nderlying \textbf{M}echanisms), a novel diagnostic framework designed to capture scale-dependent coordination between internal pathways utilizing four mechanistic scores to measure these signals; (2) we demonstrate that by analyzing the citation signature, FACTUM significantly outperforms state-of-the-art baselines for citation hallucination detection in long-form RAG; and (3) we provide empirical evidence challenging the conventional view, showing that strong FFN Pathway activity is constructive, not detrimental, to correct citation generation.


\section{Related Work}
\label{sec:related_work}

\subsubsection{Hallucination Detection in RAG.}\label{subsec:hallucinationdetectioninrag}
Methods for detecting hallucinations in RAG largely fall into two categories: those that treat the model as a black box, and those that analyze its internal states. Black-box approaches evaluate the model's final output without considering its internal processing \cite{llmcheck}. Techniques include using an external LLM judge \cite{saadfalcon2024aresautomatedevaluationframework,esetal2024ragas}, fine-tuning a separate detector model \cite{niuetal2024ragtruth,songetal2024rag,ravi2024lynxopensourcehallucination,kovács2025lettucedetecthallucinationdetectionframework}, or checking for consistency across sampled outputs \cite{manakul2023selfcheckgptzeroresourceblackboxhallucination}. The primary limitation of these approaches is that they are purely behavioral \cite{ji2024llminternalstatesreveal,su2024unsupervisedrealtimehallucinationdetection}. They can flag a potential hallucination, but they cannot reveal the underlying mechanistic signals within the model that give insight into the failure.
In contrast, methods that analyze model internals seek measurable signals of hallucination \cite{orgad2025llmsknowshowintrinsic}. While early approaches focused on general uncertainty, leveraging signals such as token probabilities and entropy \cite{logprobsuncertainty,kadavath2022languagemodelsmostlyknow,malinin2021uncertaintyestimationautoregressivestructured}, the most relevant prior work is ReDeEP which introduced a framework for decoupling a model's reliance on external context versus parametric knowledge \cite{sun2025redeepdetectinghallucinationretrievalaugmented}. However, ReDeEP treats these pathways as independent, monolithic entities and relies on imprecise proxies that significantly limit its effectiveness for detecting citation hallucination.

Specifically, ReDeEP's External Context Score (ECS) considers the entire prompt, diluting the context signal from source documents with noise from system instructions and user queries. This design also conflates attention to actual evidence with attention to the functional ``attention sink'' (beginning-of-sentence token), preventing role isolation. Furthermore, its Parametric Knowledge Score (PKS) relies on LogitLens, a vocabulary-space projection that is an unreliable proxy for the residual stream's true state \cite{belrose2023elicitinglatentpredictionstransformers}. FACTUM overcomes these limitations by introducing more direct, targeted scores for each pathway. Beyond mere decoupling, we move toward a coordination-based framework that introduces the role of the attention sink (BAS) and the first direct measure of geometric alignment between pathways (PAS). This allows FACTUM to identify the scale-dependent coordination failures that monolithic baselines miss.

\subsubsection{Citation Hallucination.} \label{subsec:citationhallucination}
While hallucination in RAG is a broad topic, citation hallucination is a specific and critical sub-problem \cite{ding2025citationstrustllmgenerated}. Recent work has established its prevalence, confirming that even state-of-the-art models frequently misattribute or fabricate sources \cite{magesh2024hallucinationfreeassessingreliabilityleading}. Existing studies have approached this from a black-box perspective, for instance by asking an LLM follow-up questions to test source comprehension \cite{agrawal2024languagemodelsknowtheyre,ni2025fullyexploitingllminternal}. However, this fails to address the core mechanistic question: Is there a measurable signature within the model's internal information pathways that reliably distinguishes a correct citation from a hallucinated citation? Our work is the first to characterize this internal signature, linking citation hallucination to predictable patterns in the model's internal states.

\section{The FACTUM Framework}
\label{sec:factum_framework}

\begin{figure}[ht!]
    \centering
    \includegraphics[width=0.82\textwidth]{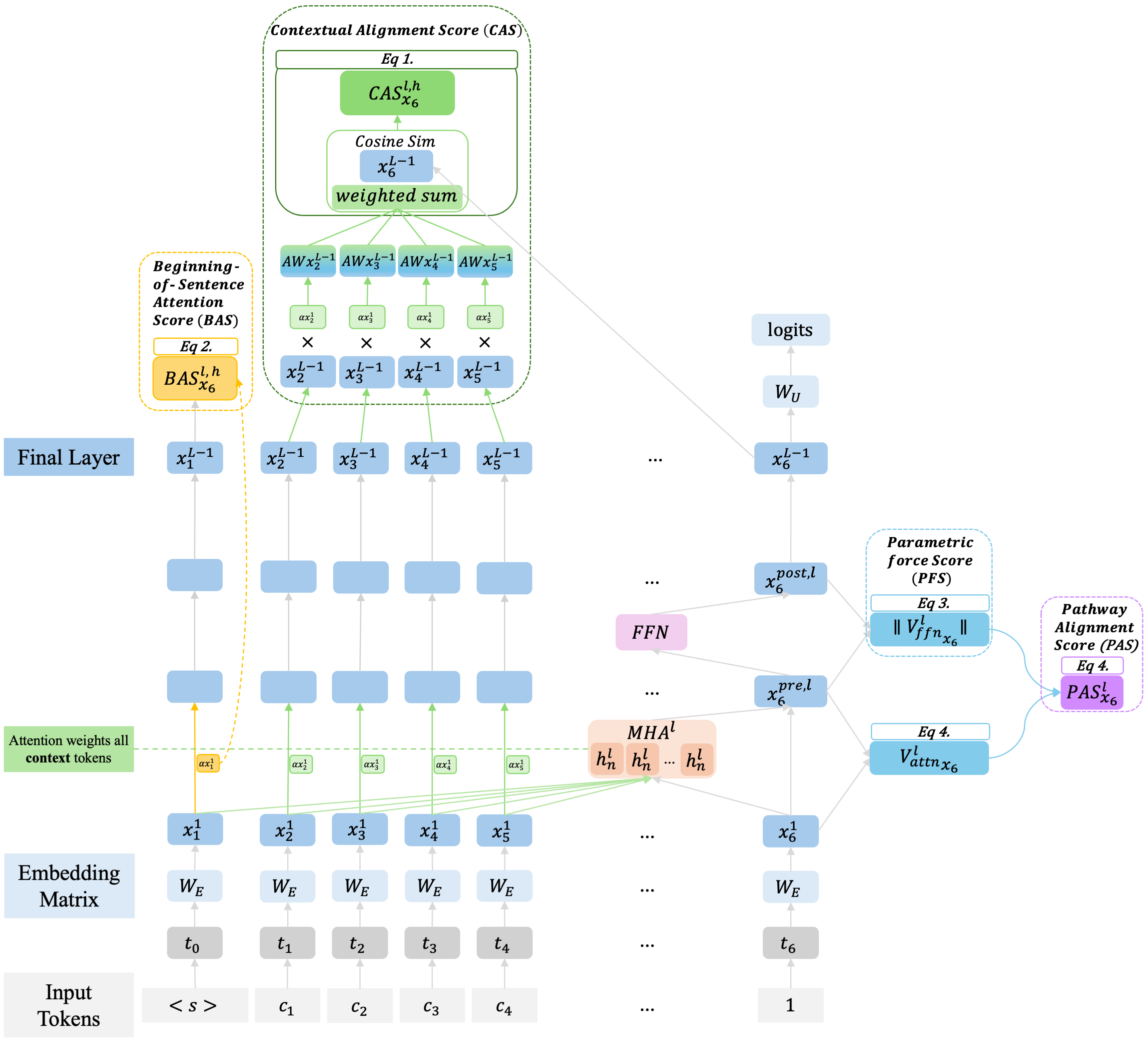} 
    \footnotesize
    \caption{The FACTUM framework analyzing internal states at a citation token (e.g.,~\texttt{1}), building on \cite{sun2025redeepdetectinghallucinationretrievalaugmented}. It derives scores from two pathways: from the Attention Pathway, it computes the Contextual Alignment Score (CAS) to measure grounding in the source documents, and the Beginning-of-Sentence Attention Score (BAS) to measure information synthesis; from the FFN pathway it measures Parametric Force Score (PFS) to quantify the magnitude of the FFN's update, and the Pathway Alignment Score (PAS) to assess the geometric alignment between the two pathway updates.}
    \label{fig:factum}
\end{figure}

The FACTUM framework leverages the Transformer architecture to provide a direct, mechanistic view into the model's internal states during citation generation. This approach is grounded in the mechanistic interpretability view of Transformers, treating hidden states as points in a structured, high-dimensional residual stream $\mathbf{x} \in \mathbb{R}^d$ \cite{elhage2021mathematical}. Following prior work, we treat this stream as a shared communication channel where each sub-layer reads from the preceding state and performs an additive write to update it  \cite{Hussain,mak2025residual}. For a citation token at position $i$, the state at layer $l$ is defined as the sum of the previous state and the updates from the Multi-Head Attention (MHA) and Feed-Forward Network (FFN) sub-layers:
\begin{equation}
    \mathbf{x}_i^{(l)} = \mathbf{x}_i^{(l-1)} + \mathbf{v}_{\text{attn}, i}^{(l)} + \mathbf{v}_{\text{ffn}, i}^{(l)}
\end{equation}
This linear, additive structure allows us to move beyond black-box analysis by decomposing internal processing into two primary information pathways: the \textbf{Attention pathway} (\textit{read}) and the \textbf{FFN pathway} (\textit{recall}). By measuring their individual contributions to the communication channel, we define four mechanistic scores designed to provide more direct, robust, and targeted measurements than prior methods. Together, these scores allow us to test our central hypothesis that citation hallucination arises from a systemic failure in the coordination between contextual grounding, internal synthesis, parametric force, and pathway alignment. Figure~\ref{fig:factum} provides a visual representation of these scores and how they are calculated.

\subsection{The Attention Pathway}
\label{sec:attn}
The Attention Pathway acts as the model's \textit{reader} and is the sole mechanism for moving information between token positions \cite{Hussain,Vaswani}. In RAG, its purpose is to contextualize the current token by selectively retrieving information from the source documents $T_C$ \cite{lietal2019information,daietal2022knowledge}. Mechanistically, this pathway performs targeted feature transport: it uses the QK circuit to determine the attention pattern $A_{i,j}$ (where to look) and the OV circuit ($W_{OV}$) to extract specific semantic content (what to read) \cite{elhage2021mathematical}. The resulting update $\mathbf{v}_{\text{attn}}$ aggregates this retrieved evidence into the residual stream:
\begin{equation}
    \mathbf{v}_{\text{attn}, i}^{(l)} = \sum_{j \in T_C} A_{i,j}^{(h,l)} (\mathbf{x}_j^{(l-1)} W_{OV}^{(h,l)})
\end{equation}
 
We introduce two scores to quantify how this reading mechanism is utilized to ground and synthesize information retrieved from the source documents.

\subsubsection{Context Alignment Score (CAS).}
To overcome the signal dilution of prior methods that consider the entire prompt, CAS provides a precise measure of contextual grounding by focusing exclusively on the source document tokens ($T_C$). It computes an attention-weighted average of the final-layer hidden states ($\mathbf{h}^{(L)}$) of all context tokens to form a holistic context vector, reflecting the model's representation of the evidence. This weighting by attention ensures that the most attended-to context tokens have the greatest influence. The score is the cosine similarity between this context vector and the generated citation token's own final-layer hidden state:
\begin{equation}
    \text{CAS}^{(h,l)}(t_i) = \frac{(\sum_{t_j \in T_C} A_{i,j}^{(h,l)} \mathbf{h}_{t_j}^{(L)}) \cdot \mathbf{h}_{t_i}^{(L)}}{||\sum_{t_j \in T_C} A_{i,j}^{(h,l)} \mathbf{h}_{t_j}^{(L)}||_2 \cdot ||\mathbf{h}_{t_i}^{(L)}||_2}
\end{equation}
where $\mathbf{h}_{t_j}^{(L)}$ represents the final-layer hidden state of context token $j$, and $A_{i,j}^{(h,l)}$ is the attention weight from head $h$ in layer $l$. A high CAS indicates strong semantic alignment between the generated token and the attended-to evidence, serving as a direct measure of contextual grounding.

\subsubsection{Beginning-of-Sentence Attention Score (BAS).}
Recent work has identified the beginning-of-sentence token \texttt{<s>} as a functional ``attention sink'' utilized for information synthesis, particularly in long contexts \cite{gu2025attentionsinkemergeslanguage}. This sink manages information flow by avoiding representational collapse and over-squashing over long contexts before making a final prediction \cite{barbero2024transformersneedglassesinformation,barbero2025llmsattendtoken}. Therefore, we hypothesize that this mechanism is used constructively to integrate retrieved information before generating a citation.
To measure this, BAS calculates the proportion of attention each head $h$ in layer $l$ allocates to the \texttt{<s>} token when generating the token citation $t_i$ at sequence position $i$:
\begin{equation}
    \text{BAS}^{(l,h)}(t_i) = A_{i,1}^{(l,h)}
    \label{eq:3}
\end{equation}
where $A_{i,1}^{(l,h)}$ is the attention weight from head $h$ in layer $l$ allocated to the first token (\texttt{<s>}) when generating citation $t_i$. A higher BAS suggests greater reliance on this internal synthesis mechanism, a signal entirely missed by previous approaches. For models without such a token, it measures attention to the very first token of the input, which serves the same functional role.

\subsection{The FFN Pathway}
\label{sec:ffn}
The FFN Pathway acts as the \textit{recall} mechanism, the primary channel for injecting parametric knowledge into the residual stream \cite{huang2025parammutesuppressingknowledgecriticalffns,daietal2022knowledge}. FFNs function as key-value memories where factual associations are stored and retrieved \cite{gevaetal2021transformer}. Following the key-value model, the FFN update vector at layer $l$ is defined as:
\begin{equation}
    \mathbf{v}_{\text{ffn}, i}^{(l)} = \sigma(\mathbf{x}_i^{(l)} W_{in}^{(l)}) W_{out}^{(l)}
\end{equation}
where $W_{in}$ acts as keys that detect semantic patterns in the communication channel and $W_{out}$ acts as values that inject the associated parametric information \cite{gevaetal2021transformer}. Crucially, direct interventions on FFNs have been shown to manipulate the factual content of a model's output \cite{NEURIPS20226f1d43d5,meng2023masseditingmemorytransformer,li2024pmetprecisemodelediting}. Measuring this activity allows us to quantify the parametric force behind each generated citation.

\subsubsection{Parametric Force Score.}
To provide a more direct measure of parametric influence than unreliable proxies such as LogitLens \cite{belrose2023elicitinglatentpredictionstransformers}, we measure the FFN's effect on the residual stream. Since the FFN pathway applies the model's stored factual knowledge \cite{gevaetal2021transformer,meng2023masseditingmemorytransformer}, their activity during citation generation can be interpreted as a form of internal fact-checking. Therefore, we hypothesize that a high PFS reflects the model confidently corroborating contextual evidence with its parametric knowledge. At each layer $l$, the FFN contributes an update to the residual stream. We isolate this contribution by taking the difference between the residual stream's state before the FFN ($\mathbf{x}_{\text{pre-ffn}}^{(l)}$) and immediately after the FFN block ($\mathbf{x}_{\text{post-ffn}}^{(l)}$), which we call the \textit{FFN update vector} ($\mathbf{v}_{\text{ffn}}^{(l)}$). The PFS is the L2 norm of this vector:
\begin{equation}
    \text{PFS}^{(l)} = ||\mathbf{v}_{\text{ffn}}^{(l)}||_2 = \sqrt{\sum_{k=1}^{d} (v_k^{(l)})^2}, \quad \text{where} \quad \mathbf{v}_{\text{ffn}}^{(l)} = \mathbf{x}_{\text{post-ffn}}^{(l)} - \mathbf{x}_{\text{pre-ffn}}^{(l)}
\end{equation}
Here, $||\cdot||_2$ denotes the L2 norm, and $\mathbf{x}_{\text{pre-ffn}}^{(l)}, \mathbf{x}_{\text{post-ffn}}^{(l)}$ represent the states of the residual stream before and after the FFN block. A larger norm signifies a more substantial modification to the residual stream, indicating a stronger ``force'' from the model's stored parametric knowledge.

\subsection{Pathway Interaction}
Central to our hypothesis is that citation hallucination stems from a systemic failure in coordination, measured via the geometric alignment between information pathways. While not the sole indicator, this alignment provides a rich signal of whether the model's parametric recall reinforces or contradicts the contextual evidence. To capture this interplay, we introduce the Pathway Alignment Score (PAS), the first metric designed to explicitly quantify this interaction.

\subsubsection{Pathway Alignment Score (PAS).}
The PAS metric quantifies the geometric alignment between the attention and FFN pathways by computing the cosine similarity between their respective update vectors. This provides a rich signal that distinguishes between collaboration, conflict, and uncoordinated activity. The score is calculated as:
\begin{equation}
    \text{PAS}^{(l)} = \frac{\mathbf{v}_{\text{attn}}^{(l)} \cdot \mathbf{v}_{\text{ffn}}^{(l)}}{||\mathbf{v}_{\text{attn}}^{(l)}||_2 \cdot ||\mathbf{v}_{\text{ffn}}^{(l)}||_2}, \quad \text{where} \quad \mathbf{v}_{\text{attn}}^{(l)} = \mathbf{x}_{\text{pre-ffn}}^{(l)} - \mathbf{x}_{\text{input}}^{(l)} 
\end{equation}
where $\mathbf{v}_{\text{attn}}^{(l)}$ is the \textit{attention update vector}, which is the update from the attention block to the residual stream, calculated as the difference between the state after the attention block ($\mathbf{x}_{\text{pre-ffn}}$) and the state before it ($\mathbf{x}_{\text{input}}$). 
Scores near $+1$ indicate alignment (collaboration), scores near $0$ indicate orthogonality (unrelated processing), and scores near $-1$ indicate anti-alignment where the FFN actively negates the contextual information.

\section{Experimental Setup}
\label{sec:experimental_setup}

\subsection{Task Definition: Single-Citation Long-Form RAG}
We focus on a common and critical RAG use case: generating a long-form report where statements are grounded in retrieved sources. More formally, given an input query $q$ and a set of retrieved source documents $D = \{d_1, d_2, \ldots, d_m\}$, the model is tasked with generating a response composed of sentences with single inline citation $c_j$ pointing to a document $d_j \in D$. The generated text follows the format: ``\textit{The standard dose for adults is 100 mg daily [Source: 1].}''. We define a citation hallucination as any citation not attested by its referenced document. To achieve maximum precision, we formulate this as a token-level binary classification task on the citation digit (e.g., the digit `1'), as this token forms the critical link to the source. While focused, this token-level approach is flexible and can be aggregated to evaluate multi-source citations.

\subsection{Dataset}
Our work focuses on detecting citation hallucination in a challenging long-form, multi-document RAG scenario. To this end, we use the TREC NeuCLIR 2024 dataset\footnote{Dataset available at: https://neuclir.github.io/} \cite{neuclir2024}. Its report generation task, requiring synthesizing information from retrieved documents, is significantly more challenging than the single-passage QA tasks often used in prior RAG hallucination studies. In our experiments on NeuCLIR, we use up to 15 documents as input to the generator. This multi-document scenario creates a high risk of the ``attributional drift'' that our work aims to detect. Although NeuCLIR is a cross-lingual track, our analysis is performed in a monolingual English setting: we provide the models with machine-translated English versions of the source documents to generate English reports. Crucially, while our experimental setup uses this 15-document implementation, the FACTUM framework is agnostic to the retrieval system. It operates at the moment of generation, making its scores applicable to any RAG system, regardless of the number or quality of retrieved documents.

\subsection{Ground-Truth Labeling}
Because our mechanistic approach requires analyzing the model's unique generation trace, we must label the model's own output. Given this constraint and the need for a scalable solution, we implemented an automated LLM-as-a-judge approach using \texttt{Llama-3.1-70B-Instruct}. Recognizing that the validity of LLM-based labels is crucial, we implemented a rigorous, multi-stage validation process. First, we constrained the judge to narrow attestation judgments using the ARGUE framework \cite{argue} rather than open-ended evaluation. Second, stability testing across multiple temperatures showed highly consistent labels ($< 1.5\%$ variation). Finally, we conducted a human validation study on 100 randomly sampled judgments (50 per model) using two independent assessors. Assessors achieved substantial agreement (Avg.~$\kappa = 0.71$), and the LLM judge aligned highly with human ground truth, achieving a mean $\kappa = 0.68$ (0.64 for 8B and 0.71 for 3B). These results confirm the LLM judge's reliability as a proxy for human judgment, providing a robust foundation for our analysis.


\subsection{Models and Data Statistics}
We select two models with 128k context windows to evaluate FACTUM in long-form settings: \texttt{Llama-3.2-3B-Instruct} and \texttt{Llama-3.1-8B-Instruct}. The 8B model generated longer reports on average ($\approx$ 1,091 tokens) than the 3B model ($\approx$ 782 tokens). For the 3B model, our final dataset contains 208 correct and 93 hallucinated citation tokens. For the 8B model, we identified 465 correct and 107 hallucinated citation tokens. To train our classifiers, we use a balanced training set and evaluate on a held-out, imbalanced test set that mirrors the real-world distribution of correct vs. hallucinated citations.

\subsection{Feature Engineering and Classification}
The raw mechanistic scores from our framework produce high-dimensional feature vectors per citation token. For the 32-layer, 32-head Llama-3.1-8B model, a single score such as CAS generates 1,024 distinct measurements (one for each head at each layer) for every citation token we analyze. To mitigate overfitting risks from such high dimensionality relative to sample size, we developed a multi-stage pipeline to distill these signals into a concise feature set.

First, we prune model components, because we hypothesize that citation attribution may be localized to specialized layers and heads. Therefore, we rank all model components (i.e., each head at each layer) for each mechanistic score based on their individual correlation with the citation hallucination label using only the training set to prevent leakage. We treat the pruning percentage, $k$, as a hyperparameter to select the top-$k\%$ components independently for each mechanistic score. This approach is agnostic to architectural location, allowing us to discover whether predictive signals are distributed broadly or localized to a specialized subset. Our optimization revealed that the optimal $k$ varies by model size, suggesting significant specialization at scale; we report final results using the optimal $k$ for each model ($k=100\%$ for 3B, $k=25\%$ for 8B). 

Second, the pruned features undergo dimensionality reduction. For attention-based scores, we aggregate across the head dimension at each layer (e.g., mean, standard deviation), as heads within a layer function as an unordered set. This results in a per-layer feature vector for each score. To capture trends across the model's hierarchical layers, we engineer features utilizing sequence-aware functions (linear slope, FFT) common in time-series analysis \cite{fft}. From this candidate pool, we select the most predictive feature for each mechanistic score using in-distribution training data. This pipeline, applied identically to our FACTUM scores and the ReDeEP baselines, yields a concise final feature set; for example, our complete FACTUM model uses just four features.

Finally, we train interpretable classifiers, including Logistic Regression \cite{jurafskyspeech2025}, LightGBM \cite{NIPS20176449f44a}, and an Explainable Boosting Machine (EBM) \cite{Lou2013AccurateIM} using 10-fold stratified cross-validation. To prevent data leakage, all splits are performed at the report level, ensuring all citations from a given report belong exclusively to either the training or the testing set within any fold.

\subsection{Baselines}
To evaluate FACTUM, we benchmark its performance against state-of-the-art baselines that also rely exclusively on internal model signals\footnote{Code 
available at https://github.com/Maximeswd/citation-hallucination}. Our primary comparison is a re-implementation of the mechanistic scores from ReDeEP: the External Context Score (ECS) and the Parametric Knowledge Score (PKS) \cite{sun2025redeepdetectinghallucinationretrievalaugmented}. This allows for a direct, head-to-head comparison of feature effectiveness.
We also compare against a range of established confidence-based scores that probe model internals for uncertainty signals. This category includes token-level Perplexity \cite{huang2025repplrecalibratingperplexityuncertainty}, LN-Entropy \cite{malinin2021uncertaintyestimationautoregressivestructured}, the Energy score \cite{energy}, and the P(True) score \cite{kadavath2022languagemodelsmostlyknow}.


\section{Results and Analysis}
\label{sec:results}
In this section, we first establish the superior performance of the FACTUM framework as a practical, state-of-the-art citation hallucination detector. We then leverage FACTUM's diagnostic scores as mechanistic probes to uncover the underlying signatures of correct and hallucinated citations.

\subsection{FACTUM Establishes a New State-of-the-Art in Citation Hallucination Detection}
As shown in Table \ref{tab:neuclir_results}, our results demonstrate that FACTUM significantly improves feature-based citation hallucination detection. While some classifier-free baselines achieve competitive AUC or recall, poor precision makes them unreliable in practice. For example, P(True) achieves high recall on the 8B model, but incorrectly flags numerous correct citations as hallucinations (a high false positive rate), undermining user trust. This highlights the need for a balanced and nuanced classification approach that can reliably distinguish between correct and hallucinated citations without overwhelming the user with false alarms. 

Among the classifier-based methods, our analysis reveals a clear hierarchy: FACTUM (Ours) $>$ CAS + PFS (Ours) $>$ ECS + PKS (Baseline). Firstly, our refined CAS + PFS (Ours) significantly outperforms the ReDeEP baseline. This confirms the superior signal quality of direct pathway scores that avoid the noise of LogitLens and prompt-level attention dilution. Second, the full FACTUM framework achieves the highest overall performance, winning 27 of 30 head-to-head comparisons with a peak AUC of 0.737 with the Logistic Regression classifier. Crucially, FACTUM increases 8B model precision from 0.201 to 0.334, which is a 66\% relative improvement over the ReDeEP baseline. This statistically significant improvement of the full FACTUM framework over the already strong ``CAS + PFS (Ours)'' (indicated by the ${\dagger}$ symbol), validates our central hypothesis: that a complete and reliable detection framework must account not only for the primary information pathways but also for their interaction and synthesis, signals uniquely captured by our PAS and BAS scores.

\begin{table}[h!]
\centering
\caption{Mean performance on the NeuCLIR 2024 dataset, averaged over 10-fold stratified cross-validation. The table is split into two sections: classifier-free \textit{General Baselines} and our main classifier-based results. In the bottom section, highlighting is scoped within each classifier block, with the best (\best{bold}) and runner-up (\runnerup{underlined}) performance shown. A star ($^{\star}$) indicates a statistically significant improvement over `ECS+PKS (Baseline)' ($p_{\text{FDR}} < 0.05$) using a two-tailed t-test with Benjamini-Hochberg (BH) correction. A dagger (${\dagger}$) indicates that the improvement of `FACTUM (Ours)' over `CAS+PFS (Ours)' is statistically significant.}
\label{tab:neuclir_results}
\resizebox{\textwidth}{!}{%
\begin{tabular}{@{}llcccccccccc@{}}
\toprule
\multirow{2}{*}{\textbf{Classifier}} & \multirow{2}{*}{\textbf{Method}} & \multicolumn{5}{c}{\textbf{Llama-3.2-3B-Instruct}} & \multicolumn{5}{c}{\textbf{Llama-3.1-8B-Instruct}} \\
\cmidrule(lr){3-7} \cmidrule(lr){8-12}
& & \textbf{AUC} & \textbf{PCC} & \textbf{Precision} & \textbf{Recall} & \textbf{F1} & \textbf{AUC} & \textbf{PCC} & \textbf{Precision} & \textbf{Recall} & \textbf{F1} \\
\midrule
\multirow{4}{*}{\textit{\parbox{1.5cm}{\centering General Baselines}}} 
 & \quad Perplexity &  0.638 & 0.085 & 0.373 &  {0.739} &  {0.489} & 0.670 & 0.137 &  {0.237} & 0.629 &  {0.334} \\
 & \quad LN-Entropy & 0.637 &  {0.189} &  {0.376} & 0.717 & 0.487 &  {0.671} &  {0.219} &  {0.237} & 0.625 & 0.333 \\
 & \quad Energy & 0.606 & 0.151 & 0.373 & 0.560 & 0.437 & 0.617 & 0.140 & 0.219 & 0.419 & 0.269 \\
 & \quad P(True) & 0.481 & -0.035 & 0.278 & 0.588 & 0.369 & 0.481 & -0.019 & 0.155 &  {0.915} & 0.261 \\  
\midrule[1.0pt]
 & \quad ECS+PKS (Baseline) & 0.610 & 0.180 & 0.391 & 0.583 & 0.463 & 0.613 & 0.151 & 0.242 & 0.571 & 0.337 \\
 \rowcolor{oursbase} & \quad CAS+PFS (Ours) & \underline{0.668}$^{\star}$ & \underline{0.272}$^{\star}$ & \underline{0.429}$^{\star}$ & \underline{0.587} & \underline{0.488} & \underline{0.687}$^{\star}$ & \underline{0.255}$^{\star}$ & \underline{0.278}$^{\star}$ & \underline{0.660}$^{\star}$ & \underline{0.388}$^{\star}$ \\
 \rowcolor{oursfull} \multirow{-3}{*}{EBM} & \quad FACTUM (Ours) & \textbf{0.700}$^{\star\dagger}$ & \textbf{0.341}$^{\star\dagger}$ & \textbf{0.473}$^{\star\dagger}$ & \textbf{0.591} & \textbf{0.515}$^{\star\dagger}$ & \textbf{0.736}$^{\star\dagger}$ & \textbf{0.326}$^{\star\dagger}$ & \textbf{0.322}$^{\star\dagger}$ & \textbf{0.669}$^{\star}$ & \textbf{0.432}$^{\star\dagger}$ \\
 \cmidrule(l){2-12}
 & \quad ECS+PKS (Baseline) & 0.588 & 0.149 & 0.388 & 0.553 & 0.450 & 0.549 & 0.070 & 0.212 & 0.534 & 0.302 \\
 \rowcolor{oursbase} & \quad CAS+PFS (Ours) & \underline{0.645}$^{\star}$ & \underline{0.237}$^{\star}$ & \underline{0.426}$^{\star}$ & \underline{0.604}$^{\star}$ & \underline{0.491}$^{\star}$ & \underline{0.592}$^{\star}$ & \underline{0.124}$^{\star}$ & \underline{0.236}$^{\star}$ & \underline{0.583}$^{\star}$ & \underline{0.333}$^{\star}$ \\
 \rowcolor{oursfull} \multirow{-3}{*}{LightGBM} & \quad FACTUM (Ours) & \textbf{0.682}$^{\star\dagger}$ & \textbf{0.302}$^{\star\dagger}$ & \textbf{0.436}$^{\star}$ & \textbf{0.624}$^{\star}$ & \textbf{0.507}$^{\star}$ & \textbf{0.693}$^{\star\dagger}$ & \textbf{0.269}$^{\star\dagger}$ & \textbf{0.300}$^{\star\dagger}$ & \textbf{0.652}$^{\star\dagger}$ & \textbf{0.408}$^{\star\dagger}$ \\
 \cmidrule(l){2-12}
 & \quad ECS+PKS (Baseline) & 0.668 & 0.272 & 0.406 & 0.636 & 0.491 & 0.536 & 0.049 & 0.201 & 0.529 & 0.290 \\
 \rowcolor{oursbase} & \quad CAS+PFS (Ours) & \underline{0.676} & \underline{0.282} & \underline{0.431}$^{\star}$ & \underline{0.667}$^{\star}$ & \underline{0.519}$^{\star}$ & \underline{0.677}$^{\star}$ & \underline{0.243}$^{\star}$ & \underline{0.275}$^{\star}$ & \underline{0.637}$^{\star}$ & \underline{0.381}$^{\star}$ \\
 \rowcolor{oursfull} \multirow{-3}{*}{Log. Reg.} & \quad FACTUM (Ours) & \textbf{0.715}$^{\star\dagger}$ & \textbf{0.354}$^{\star\dagger}$ & \textbf{0.484}$^{\star\dagger}$ & \textbf{0.693}$^{\star\dagger}$ & \textbf{0.565}$^{\star\dagger}$ & \textbf{0.737}$^{\star\dagger}$ & \textbf{0.330}$^{\star\dagger}$ & \textbf{0.334}$^{\star\dagger}$ & \textbf{0.688}$^{\star\dagger}$ & \textbf{0.446}$^{\star\dagger}$ \\
\bottomrule
\end{tabular}%
}
\end{table}

\subsection{Uncovering the Mechanistic Signatures of Citation Hallucination}
Beyond detection, FACTUM serves as a diagnostic tool to uncover the internal signatures of citation hallucination. We initially hypothesized that correct citations would be characterized by higher contextual alignment, greater attention sink usage, higher parametric force, and higher pathway alignment. Table \ref{tab:mechanistic_signatures} demonstrates that the 3B model using 100\% of its model components strongly supports this: Correct citations are indeed marked by significantly higher CAS, BAS, PAS, and PFS. This indicates that at this scale, utilizing all model components, the most effective strategy for correct citation generation is one of direct reinforcement, where all internal mechanisms work in concert.

In contrast, the 8B model's signature reveals a shift toward mechanistic specialization. Its peak performance requires pruning to its top 25\% of model components, suggesting that the task is localized to a specialized set of internal mechanisms. Within this subset, while the constructive roles of high PFS and BAS remain consistent, CAS becomes inconsistent, and PAS becomes significantly lower for our best-performing classifier (Logistic Regression). This suggests that within this specialized group of model components, the pathways are not required to simply agree, but to contribute distinct, orthogonal information to the residual stream. This finding relates to \textit{post-rationalized citations} \cite{wallet}. While one might assume models only correctly cite what they already `know' (High PFS), our results show that correct citations require simultaneous grounding (CAS) and stable synthesis (BAS). In reliable RAG, parametric memory does not replace context, but actively corroborates it through orthogonal coordination.

This forces a refinement of our hypothesis: as models scale, and don't use all their model components, their strategy for correct citation appears to evolve from simple ``agreement'' to ``specialized contribution'', where our most accurate detector identifies pathways providing complementary, orthogonal information. The emergence of these scale-dependent strategies indicates that a ``one-size-fits-all'' theory of citation hallucination is insufficient.

\begin{table}[tbh]
\centering
\footnotesize
\caption{Mechanistic signatures of correct vs. hallucinated citations. Significance is determined via a one-sided Mann-Whitney U test on the `Full FACTUM (Ours)' set with Benjamini-Hochberg correction. Heatmap shades (Dark, Medium, Light) represent significance at $p < 0.001$, $p < 0.01$, and $p < 0.05$, respectively. Blue cells indicate higher scores for correct citations ($\uparrow$); Orange cells indicate lower scores ($\downarrow$).}
\label{tab:mechanistic_signatures}
\resizebox{\textwidth}{!}{%
\begin{tabular}{l|c|c|c||c|c|c}
\toprule
\multirow{2}{*}{\textbf{Mechanistic Score}} & \multicolumn{3}{c||}{\textbf{Llama-3.2-3B\_Instruct}} & \multicolumn{3}{c}{\textbf{Llama-3.1-8B\_Instruct}} \\
\cmidrule(lr){2-4} \cmidrule(lr){5-7}
 & \textbf{EBM} & \textbf{LightGBM} & \textbf{Log. Reg.} & \textbf{EBM} & \textbf{LightGBM} & \textbf{Log. Reg.} \\
\midrule
\textbf{BAS} (Attention Sink Usage) & \cellcolor{upthree}{$\uparrow$} & \cellcolor{upone}{$\uparrow$} & \cellcolor{upthree}{$\uparrow$} & \cellcolor{upthree}{$\uparrow$} & \cellcolor{upthree}{$\uparrow$} & \cellcolor{upthree}{$\uparrow$} \\
\textbf{CAS} (Contextual Alignment) & \cellcolor{upthree}{$\uparrow$} & \cellcolor{upthree}{$\uparrow$} & \cellcolor{upthree}{$\uparrow$} & \cellcolor{upthree}{$\uparrow$} & \cellcolor{downthree}{$\downarrow$} & --- \\
\textbf{PAS} (Pathway Alignment)    & \cellcolor{upthree}{$\uparrow$} & \cellcolor{upthree}{$\uparrow$} & \cellcolor{upthree}{$\uparrow$} & \cellcolor{uptwo}{$\uparrow$} & \cellcolor{uptwo}{$\uparrow$} & \cellcolor{downthree}{$\downarrow$} \\
\textbf{PFS} (Parametric Force)     & \cellcolor{upthree}{$\uparrow$} & \cellcolor{upthree}{$\uparrow$} & \cellcolor{upthree}{$\uparrow$} & \cellcolor{upthree}{$\uparrow$} & \cellcolor{upthree}{$\uparrow$} & \cellcolor{upthree}{$\uparrow$} \\
\bottomrule
\end{tabular}%
}
\end{table}

\subsection{Case Study: Mechanistic Analysis of Figure~\ref{fig:main_example}} 
We analyze Fig.~\ref{fig:main_example}’s transition through our mechanistic signals. While the correct case (Left) reflects correspondence, the hallucinated scenario (Right) reveals \textit{attributional drift}: the FFN pathway applies significant force (High PFS) toward the ``100 mg'' claim, but the Attention pathway fails to ground the citation (Low CAS). Decreased BAS further suggests a synthesis failure, indicating the citation is driven by parametric likelihood rather than verified contextual grounding. 

Correct citation requires scale-dependent \textit{functional coordination}. In the 3B model, this manifests as reinforcement (High PAS); in the 8B model, it shifts toward specialized, non-redundant contributions (Low PAS). Conversely, incorrectly citing Document 2 triggers \textit{mechanistic dissonance}. Despite conflicting context (10 mg), the FFN pathway prioritizes its internal ``100 mg'' memory (Sustained PFS). This marks a universal coordination failure where the FFN pathway operates independently, providing neither the reinforcement (3B) nor the specialization (8B) required for correctness. By identifying this dissonance, FACTUM serves as a verifiability guardrail against failures entirely invisible to black-box methods, further corroborated by concurrent mechanistic work~\cite{vandort2026}.

\subsubsection{Concluding Remarks.}
We introduced FACTUM, a framework setting a new state-of-the-art for citation hallucination detection through internal pathway analysis. Beyond detection, our work reveals that the signature of correctness is not static, but evolves with model scale. By demonstrating that parametric force acts as a constructive corroborator rather than a mere source of hallucination, we challenge prevailing assumptions and pave the way for more nuanced, mechanistic guardrails in reliable RAG systems.

\subsubsection{A Final Note on Trust of Citations.}
\label{sec:final_note}
A question for the reader: Did you notice our citation of Shakespeare in the introduction to support a claim about how citations incite trust? Although the Sonnets explore themes of trust, making the reference seem plausible, it was a deliberate false citation. Now imagine if the citation appeared as: ``...incite trust in the reader \cite{shakespeare}\textcolor{red}{*}''. That asterisk would represent a potential fix: a visual warning generated by a system like FACTUM, indicating the citation's grounding is highly suspect. We included this small deception to make the central point of this paper tangible: even for experts reading a paper on this exact topic, the pull to trust a citation is powerful. This subtle gap between plausibility and verifiable truth is precisely highlighting the critical need for our robust, mechanistic detection method FACTUM.

\subsubsection{\ackname} This research was supported by project VI.Vidi.223.166 of the NWO Talent Programme, which is (partly) financed by the Dutch Research Council (NWO). 

\subsubsection{Disclosure of Interests.} The authors have no competing interests to declare that are relevant to the content of this article.

%
%
\bibliographystyle{splncs04}
\bibliography{biblio}
\end{document}